\documentclass[prd,superscriptaddress,onecolumn,preprintnumbers,%
  amsmath,amssymb]{revtex4-2}
\usepackage[utf8]{inputenc}

\usepackage{amsmath,bm}
\usepackage{graphicx,csvsimple,booktabs,epsfig}

\date{\today}

\usepackage{verbatim}
\usepackage[breaklinks,colorlinks=true]{hyperref}
\usepackage{mathtools}
\usepackage{amsthm,amssymb}
\usepackage{lipsum}
\usepackage{calrsfs}
\allowdisplaybreaks
\usepackage[dvipsnames]{xcolor}
\newcommand\myshade{85}
\colorlet{mylinkcolor}{BrickRed}
\colorlet{mycitecolor}{NavyBlue}
\colorlet{myurlcolor}{Aquamarine}
\hypersetup{
  linkcolor  = mylinkcolor!\myshade!black,
  citecolor  = mycitecolor!\myshade!black,
  urlcolor   = myurlcolor!\myshade!black,
  colorlinks = true,
}

\begin{document}
\title{Cooperative effects in feature importance of individual patterns:  application to air  pollutants and  Alzheimer’s disease }

\author{M. Ontivero-Ortega}
\affiliation{Università degli Studi di Bari Aldo Moro, Dipartimento Interateneo di Fisica M. Merlin, Bari, 70125, Italy}
\affiliation{Cuban Center for Neuroscience, Havana, Cuba}
\affiliation{Istituto Nazionale di Fisica Nucleare, Sezione di Bari, Bari, 70125, Italy}
\author{A. Fania}
%
\affiliation{Università degli Studi di Bari Aldo Moro, Dipartimento Interateneo di Fisica M. Merlin, Bari, 70125, Italy}
\affiliation{Istituto Nazionale di Fisica Nucleare, Sezione di Bari, Bari, 70125, Italy}

\author{A. Lacalamita}

\affiliation{Università degli Studi di Bari Aldo Moro, Dipartimento Interateneo di Fisica M. Merlin, Bari, 70125, Italy}
\affiliation{Istituto Nazionale di Fisica Nucleare, Sezione di Bari, Bari, 70125, Italy}

\author{R. Bellotti}

\affiliation{Università degli Studi di Bari Aldo Moro, Dipartimento Interateneo di Fisica M. Merlin, Bari, 70125, Italy}
\affiliation{Istituto Nazionale di Fisica Nucleare, Sezione di Bari, Bari, 70125, Italy}

\author{A. Monaco}
\affiliation{Università degli Studi di Bari Aldo Moro, Dipartimento Interateneo di Fisica M. Merlin, Bari, 70125, Italy}
\affiliation{Istituto Nazionale di Fisica Nucleare, Sezione di Bari, Bari, 70125, Italy}

\affiliation{These authors contributed equally to this work and share last authorship.}
\author{S. Stramaglia}
\affiliation{Università degli Studi di Bari Aldo Moro, Dipartimento Interateneo di Fisica M. Merlin, Bari, 70125, Italy}
\affiliation{Istituto Nazionale di Fisica Nucleare, Sezione di Bari, Bari, 70125, Italy}

\affiliation{These authors contributed equally to this work and share last authorship.}

\begin{abstract}
Leveraging recent advances in the analysis of synergy and redundancy in systems of random variables, an adaptive version of the widely used metric Leave One Covariate Out (LOCO) has been recently proposed to quantify cooperative effects in feature importance (Hi-Fi), a key technique in explainable artificial intelligence (XAI), so as to disentangle high-order effects involving a particular input feature in regression problems. Differently from standard feature importance tools, where a single score measures the relevance of each feature, each feature is here characterized by three scores, a two-body (unique) score and higher-order scores (redundant and synergistic). This paper presents a framework to assign those three scores (unique, redundant, and synergistic) to  each individual pattern of the data set, while comparing it with the well-known measure of feature importance named {\it Shapley effect}. 
To illustrate the potential of the proposed framework, we focus on a One-Health application: the relation between air pollutants and Alzheimer's disease mortality rate. Our main result  is the synergistic association between features related to $O_3$ and $NO_2$ with mortality, especially in the provinces of Bergamo e Brescia; notably also the density of urban green areas displays synergistic influence with pollutants for the prediction of AD mortality. Our results place local Hi-Fi as a promising tool of wide applicability, which opens new perspectives for XAI as well as to analyze high-order relationships in complex systems.

\end{abstract}
\maketitle

\section{Introduction}
In \cite{ontivero_2025} a novel global feature importance method for regression has been introduced for explainable artificial intelligence (XAI) \cite{gunning_2019}, based on recent results which generalize the traditional dyadic description of networks of variables to the higher-order setting \cite{battiston_2021,rosas_2022}. 
Notably, an increasing attention is being devoted to the emergent properties of complex systems, with a prominent role in this literature played by partial information decomposition (PID) \cite{williams_2010} and its subsequent developments \cite{lizier_2018}, exploiting information-theoretic tools to reveal high-order dependencies among groups of three or more random variables and describe their synergistic or redundant nature \cite{barrett_exploration_2015,gatica_high-order_2021,luppi,rosas_quantifying_2019,stramaglia_quantifying_2021}. Within this framework, redundancy refers to information retrievable from multiple sources, while synergy refers to statistical relationships existing within the whole system that cannot be observed in its individual parts. 

The approach described in \cite{ontivero_2025}, named Hi-Fi (high-order interactions for feature importance), is rooted on a well known metric of feature importance named Leave-One-Out Covariates (LOCO) \cite{loco_2018}, i.e. the reduction of the prediction error when the feature under consideration is added to the set of all the features used for regression, and proposes an adaptive version of LOCO which provides three scores for each feature:  the unique pure standalone (two-body) influence of the feature on the target, and the contributions stemming from synergistic and redundant interactions with other features.
It is worth mentioning that the decomposition of feature importance in \cite{ontivero_2025} clearly depends also on the choice of the hypothesis space for regression, hence it should be assumed that a proper model for data has been selected.

In this paper we introduce  the \emph{local Hi-Fi decomposition}, whose goal is assigning to each pattern three scores so as to quantify how much the features have cooperated on each particular pattern to produce the correct or wrong output, as well as the two-way information by each feature;   ensemble averages of local quantities correspond to the global unique, redundant and synergistic scores as evaluated in \cite{ontivero_2025}.
We remark that the connection between local Hi-Fi and the corresponding global metrics is conceptually the same as those between local mutual information,  local transfer entropy and local Granger causality \cite{lte,lgc} with the corresponding local quantities.

The rest of the paper is organised as follows. Section II provides background information about the global Hi-Fi, and introduces the local Hi-Fi definition. Then, Section III illustrates the proposed method in a case study (the well-known dataset of wine quality), and Section IV presents an application to One-Health. i.e. the relation between air pollutants and Alzheimer’s disease mortality. Section V summarizes  our main conclusions.

\section{The method}
Firstly, we briefly recall the method developed in \cite{ontivero_2025} and named HiFi. Let consider $n$ stochastic variables ${\bf Z}=\{z_\alpha \}_{\alpha=1,\ldots,n}$, a driver variable $x$ and a response variable $y$. Let $\epsilon \left(y|x,{\bf Z}\right)$ be the mean squared prediction error  of $y$ on the basis of all input variables at hand, $x$ and ${\bf Z}$ (corresponding to linear regression or any non-linear regression model). 
Now, consider the prediction error
of $y$ on the basis of just the ${\bf Z}$ variables, $\epsilon \left(y |{\bf
Z}\right)$, the reduction of 
errors in the two
conditions
\begin{equation}\label{loco}
L_{\bf Z}(x\to y) =\epsilon \left(y |{\bf
Z}\right)-\epsilon \left(y|x,{\bf
Z}\right)
\end{equation}
is the well known measure of feature importance LOCO \cite{loco_2018}.
With the same notation, the reduction of error variance using only the driver $x$  (the {\it pairwise} predictive power, or  explained variance) will be denoted $L_{\bf \emptyset}(x\to y)= \sigma^2_y -\epsilon \left(y|x\right)$. 
Consider now just a subset ${\bf z}$ of all the variables in ${\bf Z}$: searching for  ${\bf z_{min}}$ minimizing $L_{\bf z}(x\to y)$  captures the amount of redundancy $R$ that the rest of variables shares with the pair $x-y$, and the redundancy score is defined $R=L_{\bf \emptyset}(x\to y)-L_{\bf z_{min}}(x\to y)$. The unique predictive power $U$, i.e. the pure two-body influence of $x$ on $y$, is given by $U=L_{\bf z_{min}}(x\to y)$. On the other hand, searching for ${\bf z_{max}}$ maximizing $L_{\bf z}(x\to y)$, leads to the amount of synergy $S$ that the {\it remaining variables} provide in terms of the increase of predictability:  $S=L_{\bf z_{max}}(x\to y) - L_{\bf \bf \emptyset}(x\to y)$. 

By construction:
\begin{equation}
L_{\bf z_{max}}(x\to y)= S+R+U,\\
\label{eq:decomposition}
\end{equation}
i.e. the maximal LOCO, related to  the interaction $x\to y$, is decomposed into the sum of a unique contribution (U), representing a pure two-body effect, and synergistic and redundant contributions that describe cooperative effects on $y$, due to $x$ and ${\bf Z}$ variables. 
Since conducting an exhaustive search for subsets ${\bf z_{min}}$ and ${\bf z_{max}}$ is unfeasible for large $n$, in \cite{ontivero_2025} a greedy search strategy has been proposed, wherein firstly one performs a search over all the z variables for the first variable to be tentatively used. Subsequently, one variable is added at a time, to the previously selected ones, to construct the set of ${\bf z}$ variables that either maximize or minimize the $L_{\bf z}$. 
The criterion for terminating the greedy search, minimizing (maximizing) the $L_{\bf z}$, is to stop when the corresponding decrease (increase) can be explained as due to chance, according to a surrogates test. 
It is worth stressing that ${\bf z_{min}}$ and ${\bf z_{max}}$ are constituted by the set of features which, alongside with the feature under consideration, are redundant and synergistic, respectively, for the  predictive model \cite{ontivero_2025}.

To introduce the local Hi-Fi decomposition, given the data-set of $N$ patterns ${\cal D}=\{x_k,{\bf Z}_k,y_k\}_{k=1}^N$, suppose that  models $f_{ {\bf
Z}}$   and $ f_{x {\bf
Z}}$ are fitted on $\cal{D}$ using input variables  ${\bf
Z}$  and $\{x,{\bf
Z}\}$ respectively.
The local LOCO  of a given pattern $\{{\it x_i}$, ${\it Z_i}$, ${\it y_i}\}$ is then defined as follows:

\begin{equation}\label{loco}
{\cal L}_{\it Z_i}({\it x_i}\to {\it y_i}) =\left( {\it y_i}-f_{ {\bf
Z}} ({\it Z_i})\right)^2 -  \left( {\it y_i}-f_{x {\bf
Z}} ({\it x_i},{\it Z_i})\right)^2.
\end{equation}
Note that LOCO can be obtained by averaging, over the data-set, the local LOCO:
\begin{equation}\label{loco1}
L_{\bf Z}(x\to y)=\frac{1}{N}\sum_{i=1}^N
{\cal L}_{\it Z_i}({\it x_i}\to {\it y_i}).
\end{equation}

Local scores of $\{{\it x_i}$, ${\it Z_i}$, ${\it y_i}\}$  are thus defined as follows:

\begin{align*}
U_i&={\cal L}_{\it z_{min |i}}({\it x_i}\to {\it y_i}),\\
R_i&={\cal L}_{\emptyset}({\it x_i}\to {\it y_i})-{\cal L}_{\it z_{min |i}}({\it x_i}\to {\it y_i}),\\
S_i&={\cal L}_{\it z_{max| i}}({\it x_i}\to {\it y_i})-{\cal L}_{\emptyset}({\it x_i}\to {\it y_i}),\\
\end{align*}
and their ensemble average coincide with global scores by construction. It is worth stressing that, differently from the ensemble averages which are strictly non negative, local scores can assume negative values in correspondence of those patterns of variables  which are mis-informative for the correct prediction; we refer to \cite{lte,lgc} for discussions about the interpretation of negative values of local information quantities.

We will compare the proposed approach with the framework for feature importance inspired by the Shapley approach of game theory.
First of all, we  stress the difference between Shapley effects\cite{shap_val} (also referred to as Shapley sensitivity index) and Shapley values as defined in \cite{shap_val} (named SHAP - SHapley Additive exPlanations). Shapley values quantify the contribution of each feature to a specific prediction made by the model, relative to the average prediction, whilst Shapley effects, a concept primarily used in Global Sensitivity Analysis (GSA), is a variance-based sensitivity measure that quantifies the total contribution of an input variable (feature) to the overall variance of the model's output, taking into account its direct effect and its interactions with other variables. The main difference is that for Shapley values a single model is fitted using the full set of covariates, whilst evaluating Shapley effects requires fitting a different model for all subsets of covariates.

Local HiFi is related to Shapley effects \cite{song_2016}, which assigns to every feature a single score whose global expression is the following:
\begin{equation}\label{shap_global}
\Phi(x)= \sum_{{\bf z}\subseteq {\bf Z}} \frac{|{\bf z}|! (n-|{\bf z}|)!}{(n+1)!} L_{\bf z}(x\to y),
\end{equation}
whilst its local expression reads
\begin{equation}\label{shap_local}
\Phi(x_i)= \sum_{{\bf z}\subseteq {\bf Z}} \frac{|{\bf z}|! (n-|{\bf z}|)!}{(n+1)!} {\cal L}_{\it z_i}({\it x_i}\to {\it y_i}).
\end{equation}
Hence Shapley effects amount to a weighted average of LOCO values over all the subsets of the other features ${\bf Z}$; on the other hand Hi-Fi picks the subsets corresponding to the minimum and maximum LOCO, as well as the reduction in variance due to  features alone, to provide three different scores measuring cooperative effects and two-body influence.

Finally, we stress that the evaluation of Shapley effects is unfeasible but for cases with few covariates, therefore here we adopted a Montecarlo approach to evaluate it \cite{goda}.

\section{Case study: wine quality}
In order to illustrate the proposed approach, we analyze the Wine Quality data set (n = 6497) obtained from the UCL Machine Learning Repository \cite{wine_dataset}, already analyzed in \cite{ontivero_2025} in terms of global scores. The goal is to predict wine quality scores (a score between 0 and 10) based on ten physicochemical properties: fixed acidity, volatile acidity, citric acid, residual sugar, chlorides, free sulfur dioxide, total sulfur dioxide, density, pH, sulphates.

Like in \cite{ontivero_2025}, after z-scoring all variables we adopt a linear regression model as a function to predict the response variable and estimate the error variance. 
Using global scores we found that the most important feature is the density, which showed the largest unique contribution and was also the most synergistic variable together with  residual sugar. Density, chlorides and volatile acidity were the most redundant variables. 

\begin{figure}[t!]
    \centering
    \includegraphics[width=1\textwidth]{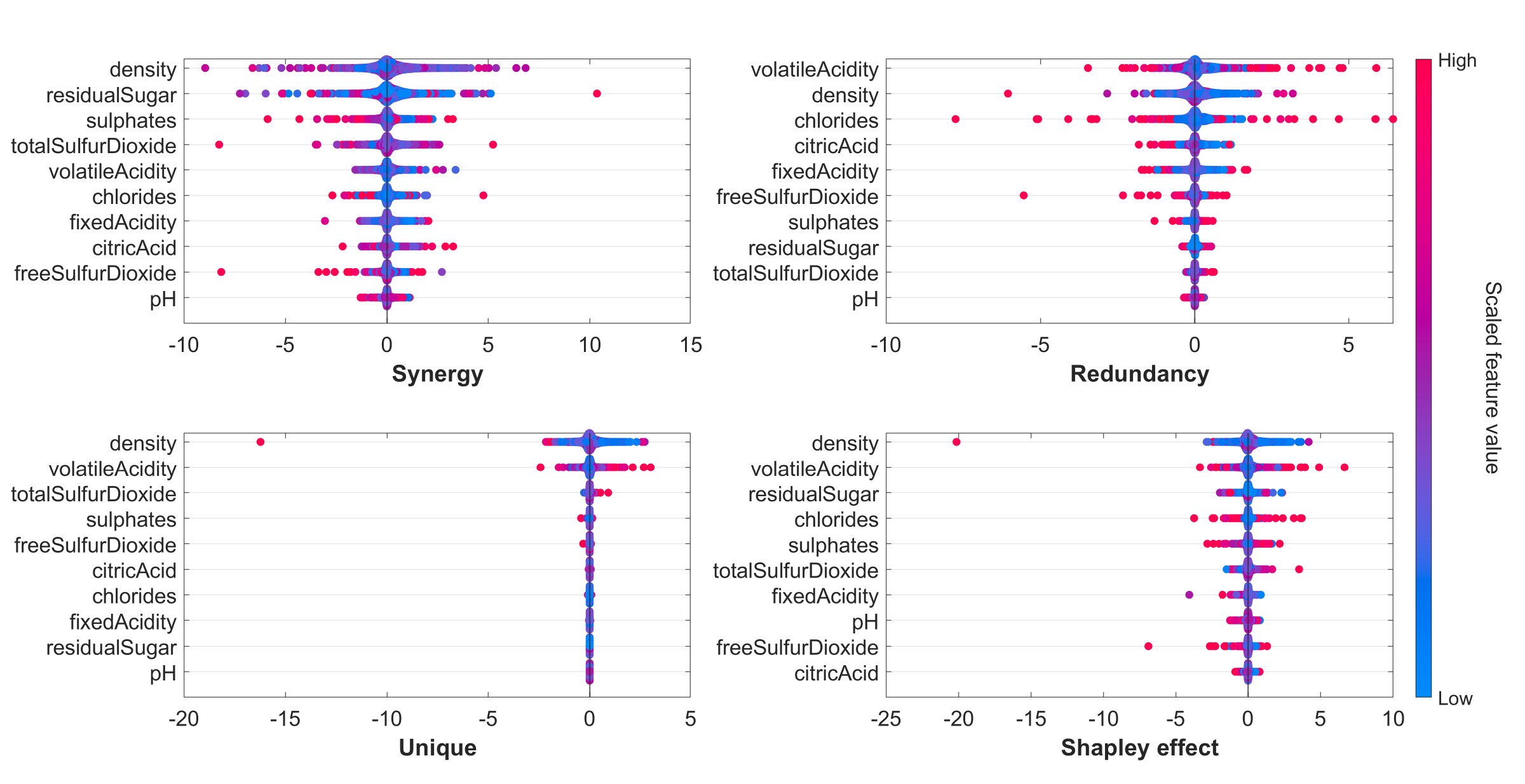}
    \caption{\textbf{Local patterns for the wine dataset. Features in each panel are ordered according to average value. Top-left panel: Local synergy of patterns. Top-right panel: Local redundancy of patterns. Bottom-left panel: Local unique scores of patterns. Bottom-right panel: Local Shapley effect.}}      
    \label{fig:wine1}
\end{figure}

In figure (\ref{fig:wine1}) we depict the local scores for the 6497 patterns, with features ordered according the corresponding global scores. The most synergistic features are density, residual sugar  and sulphates; the most redundant features are volatile acidity, density and chlorides. The features conveying unique information are density, volatile acidity and free sulfur dioxide. The Shapley effect, depicted in the bottom-right panel of (\ref{fig:wine1}), is highest for the features density, volatile acidity and residual sugar: these features are highly ranked also for $S$, $R$ and $U$, thus showing that Shapley effects condense in a single score information about the data-set which can be better captured looking separately the synergy, redundant and unique contributions.

\begin{figure}[t!]
    \centering
    \includegraphics[width=0.45\textwidth]{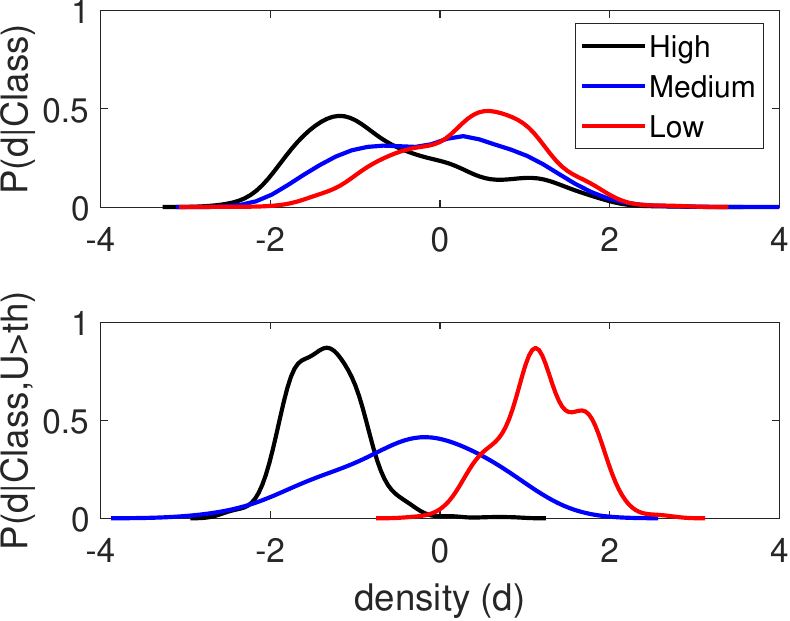}
    \caption{\textbf{ Top: The conditional probability distributions of density for the three classes of wines. Bottom:  the same quantities but also conditioned on $U> th$, where the threshold $th$ is fixed to select $15\%$ of patterns with highest U.}  }
    \label{fig:wine3}
\end{figure}

Since density is the feature with highest unique contribution, it is interesting to focus on the local values of U, for density, to stress its interpretation:  the value of $U$, on a particular pattern, measures to what extent, for that pattern, the value of the density standalone  was useful for predicting the target. A negative value means that density was mis-informative for that pattern.
To illustrate this,
we divide the patterns in three balanced groups according to the quality of wine: High, Medium and Low; in figure (\ref{fig:wine3})-top we depict the conditioned probability of density for the three classes. In figure (\ref{fig:wine3})-bottom we depict the same quantities but also conditioned on $U> th$, where $th$ has been chosen so as to select $15\%$ of patterns with highest U. We observe that selecting the highest values of U leads to better separability among classes, in particular the distributions of high and low wines appear separated: high values of $U$ select those patterns for whom the density alone is most useful to predict the class.


\section{Application: Influence of pollution, socioeconomic and health factors in Alzheimer's mortality}

In this section we apply the proposed methodology to study the relationship between Alzheimer's disease (AD) mortality and several factors from pollution, socioeconomic, and health domains. 
Currently, there is still no cure and AD is often diagnosed long time after onset because there is no clear diagnosis. Thus, it is important to investigate the risk factors that could be strongly connected to the disease onset.
In \cite{alz_data}, a publicly available data-set was analyzed using Shapley values \cite{shap_val} for feature importance in the prediction of  AD mortality, expressed as Standardized Mortality Ratio (SMR), and it was found that air pollution (mainly $O_3$ and $NO_2$) contributes the most to AD mortality prediction.

Here, we apply local HiFi to the same data as in \cite{alz_data}, with the aim of highlighting the cooperation of features, related to pollutants, to predict AD mortality. This data-set includes 32 features collected from each Italian province (107 in total) over the period 2015-2019, and grouped into five categories: Air Pollution, Soil Pollution, Urban Environment, Socioeconomic Data, and Other Pathologies (see Table 1). AD mortality was expressed as a Standardized Mortality Ratio (SMR) score, and SMR values greater than 1 indicate higher mortality incidence than the mean national value (see \cite{alz_smr} for further details). Before Hi-Fi estimation, the data for each year were combined into one large dataset (N = 535 patterns), and the entire dataset was normalized (z score for each variable). Then global and local scores were calculated using a linear regression model to predict mortality based on the 32 input features.

\begin{table}[]
\caption{Factors and Categories}\label{tab2}
\begin{tabular}{@{}ll@{}}
\toprule
Category & Indicators \\ \midrule
Air pollution (9)      & AQI, Benzene, CO, NO2, O3, PM2.5, PM10, SO2, Temperature\\
Soil pollution (6)     & Cultivated-areas, Microelement-fertilizer, N-fertilizer, Organic-fertilizer, P4O10-fertilizer, Urban-areas\\
Urban environment (7)   & Electric-consumption, urban green areas, noise, Photovoltaic-panel, urban-traffic, vehicles-total, wastes\\ 
Socioeconomic data (5) & bed-number, income, instruction, life-quality, lifetime  \\ 
Other pathologies (5) & brain-mort, circulatory-mort, diabetes-mort, digestive-mort, ischemia-mort \\ \bottomrule
\end{tabular}
\end{table}

\begin{figure}[t!]
    \centering
    \includegraphics[width=1\textwidth]{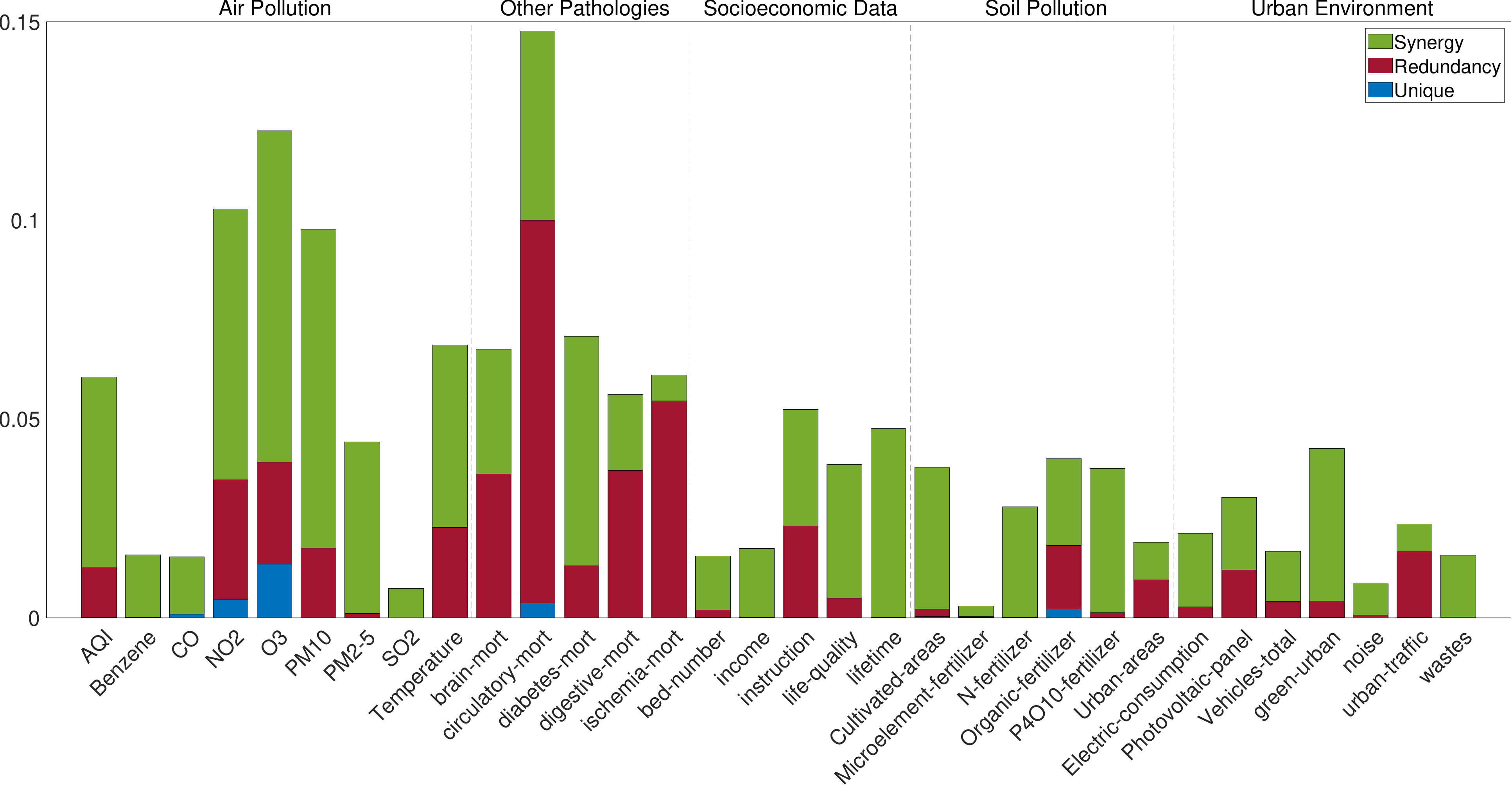}
    \caption{\textbf{Global scores of each variable for the Alzheimer dataset.}}
    \label{fig:alz_gs}
\end{figure}

The figure (\ref{fig:alz_gs}) shows the global high-order scores: air pollution features are the most synergistic group of covariates, particularly $O_3$, $PM_{10}$ and $NO_2$. Some features belonging to the group ${\it other-pathologies}$ show the highest redundancy, in particular circulatory-mort is the variables with highest total predictive power but the information it conveys is mostly redundant with those from other input variables.
This behavior may be related to a competing risk mechanism: patients affected by both cardiovascular diseases and AD often have only one of the two officially recorded as the cause of death. This competing classification can lead to overlapping information with other variables and may explain the high redundancy.

\begin{figure}[t!]
    \centering
    \includegraphics[width=.6\textwidth]{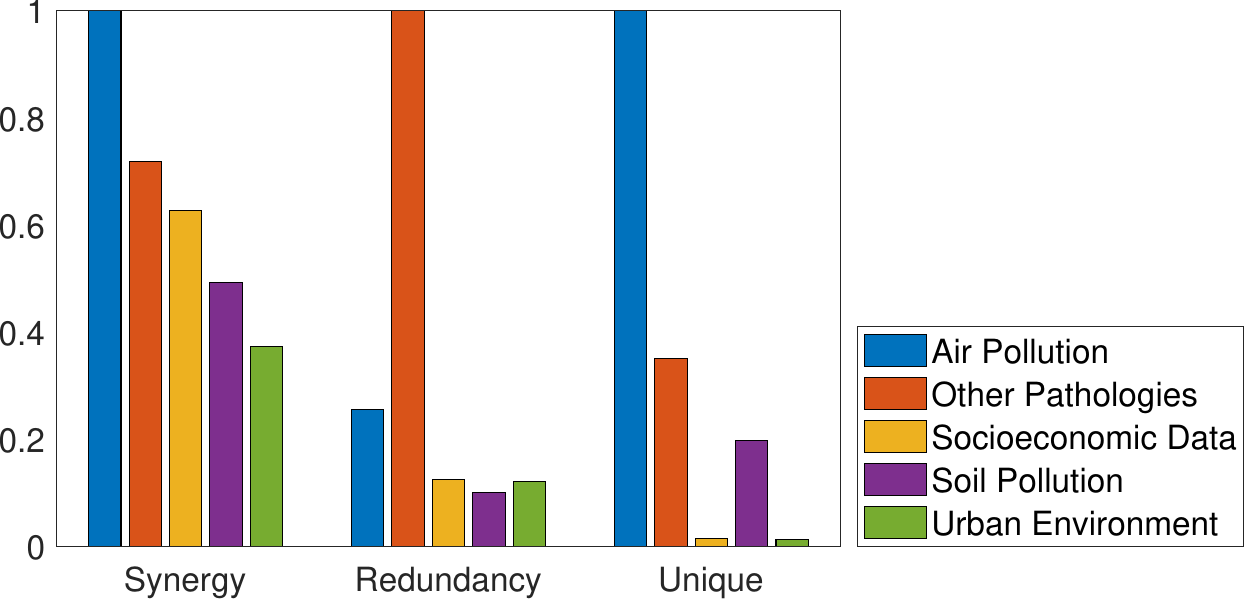}
    \caption{\textbf{Global scores of each category (averaged across all it's features) for the Alzheimer dataset. In the figure, high-order scores (Synergy, Redundancy, Unique)  were normalized so that in each category the  maximum value is one.}}
    \label{fig:alz_gs_cat}
\end{figure}

\begin{figure}[t!]
    \centering
    \includegraphics[width=.8\textwidth]{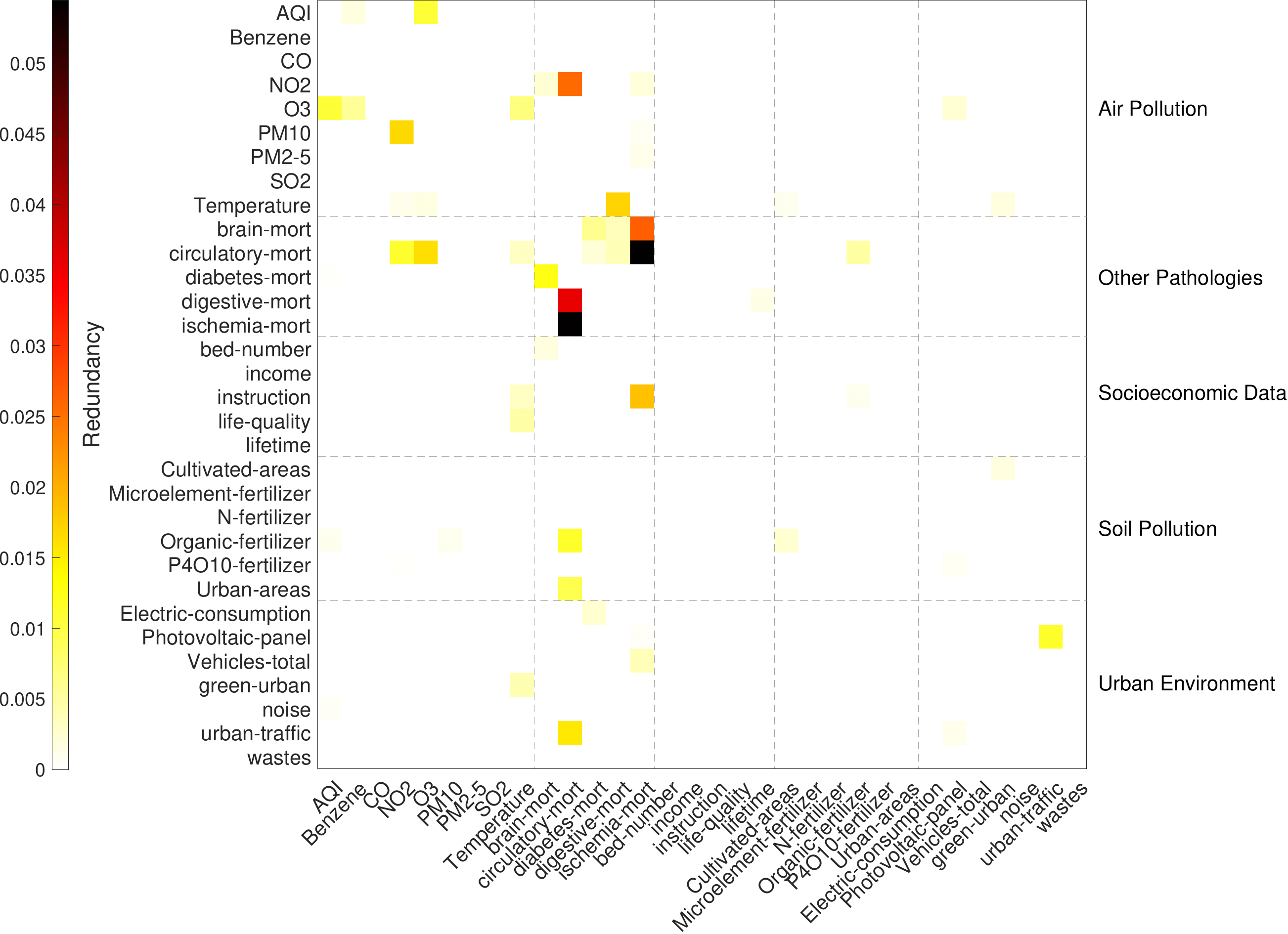}
    \caption{\textbf{The redundant multiplets for each feature of the Alzheimer dataset. The color value representing the decrement of LOCO value due to the inclusion of that variable (column) in the multiplet of a specific driver variable (row).}}
    \label{fig:alz_gs_multR}
\end{figure}

\begin{figure}[t!]
    \centering
    \includegraphics[width=.8\textwidth]{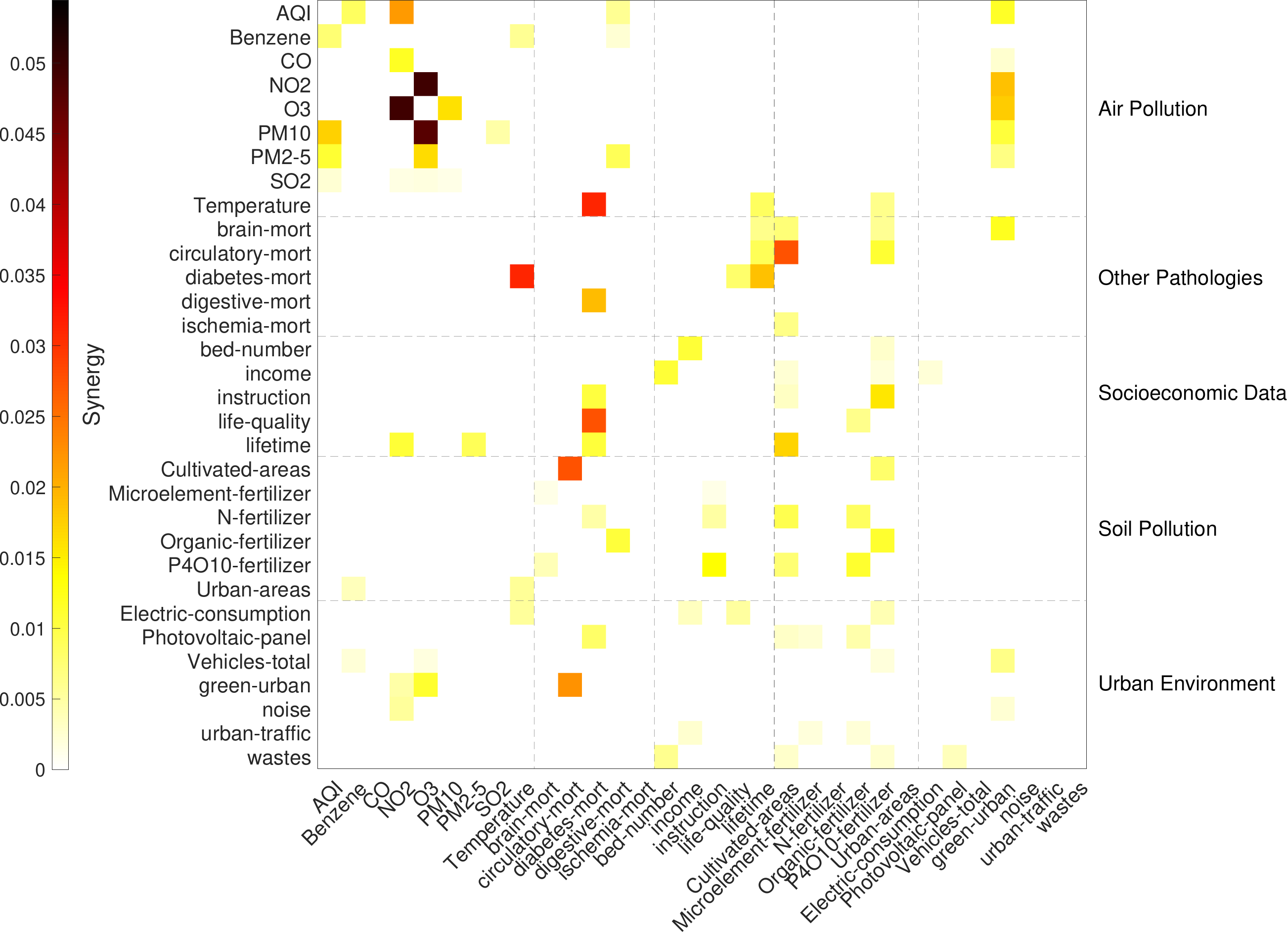}
    \caption{\textbf{The synergistic multiplets for each feature of the Alzheimer dataset. The color value representing the increment of LOCO value due to the inclusion of that variable (column) in the multiplet of a specific driver variable (row)}.}
    \label{fig:alz_gs_multS}
\end{figure}

Concerning unique scores, the features with highest two-body predictive power are, in order $O_3$, $NO_2$, and circulatory-mort. Figure (\ref{fig:alz_gs_cat}) summarizes these results, showing that Air pollution is the category that provides more unique and synergistic information, and Other pathologies category is mostly redundant. All other categories are less important or almost irrelevant, although most of their features have moderated synergistic contributions.
In figure (\ref{fig:alz_gs_multR}), for each driving feature, we depict the corresponding set of redundant variables, with a color code representing the decrement of LOCO due to the inclusion of each variable in the redundant multiplet; notably redundancy is high among features related to other pathologies, i.e. redundancy appears to be connected with co-morbidity. On the other hand, in figure (\ref{fig:alz_gs_multS}), for each driving feature, we depict the corresponding set of synergistic variables, with a color code representing the increment of LOCO due to the inclusion of each variable in the synergistic multiplet; we observe a remarkable synergy of air pollution variables, in particular $NO_2$ and $PM_{10}$ which synergistically cooperate with $O_3$.

Although $NO_2$ and $O_3$ are chemically connected through photochemical reactions, their observed synergy may arise from both their distinct spatiotemporal behavior and their complementary biological effects. Although $NO_2$ concentrations are related to direct emission sources and show strong urban variability, $O_3$ forms secondarily and exhibits different temporal dynamics \cite{han2011analysis}. Furthermore, the two pollutants affect different physiological pathways: while $NO_2$ may primarily contribute to vascular inflammation and BBB disruption, $O_3$ can further amplify oxidative neuronal injury \cite{block2009air}. This suggests that, rather than being redundant, they jointly improve predictability through cooperative mechanisms.
Moreover, we would like to stress that our results show that the density of urban green areas appears to have a cooperative synergistic action with several air pollution features.

This result may suggest a mitigating effect of urban green space on both exposure to air pollution and its health impacts \cite{diener2021can}. This behavior is particularly evident in provinces with higher green space density, such as Sondrio and Trento. Despite showing high levels of atmospheric pollutants, these northern Italian provinces report relatively low average AD mortality rates (SMR = 0.81 and 0.68, respectively). Although the green urban variable shows a weak negative correlation with SMR (–0.07), it exhibits a notable synergistic interaction with air pollution variables, indicating that green infrastructure may modulate the effects of pollutants rather than acting independently.

\begin{figure}[t!]
    \centering
    \includegraphics[width=.95\textwidth]{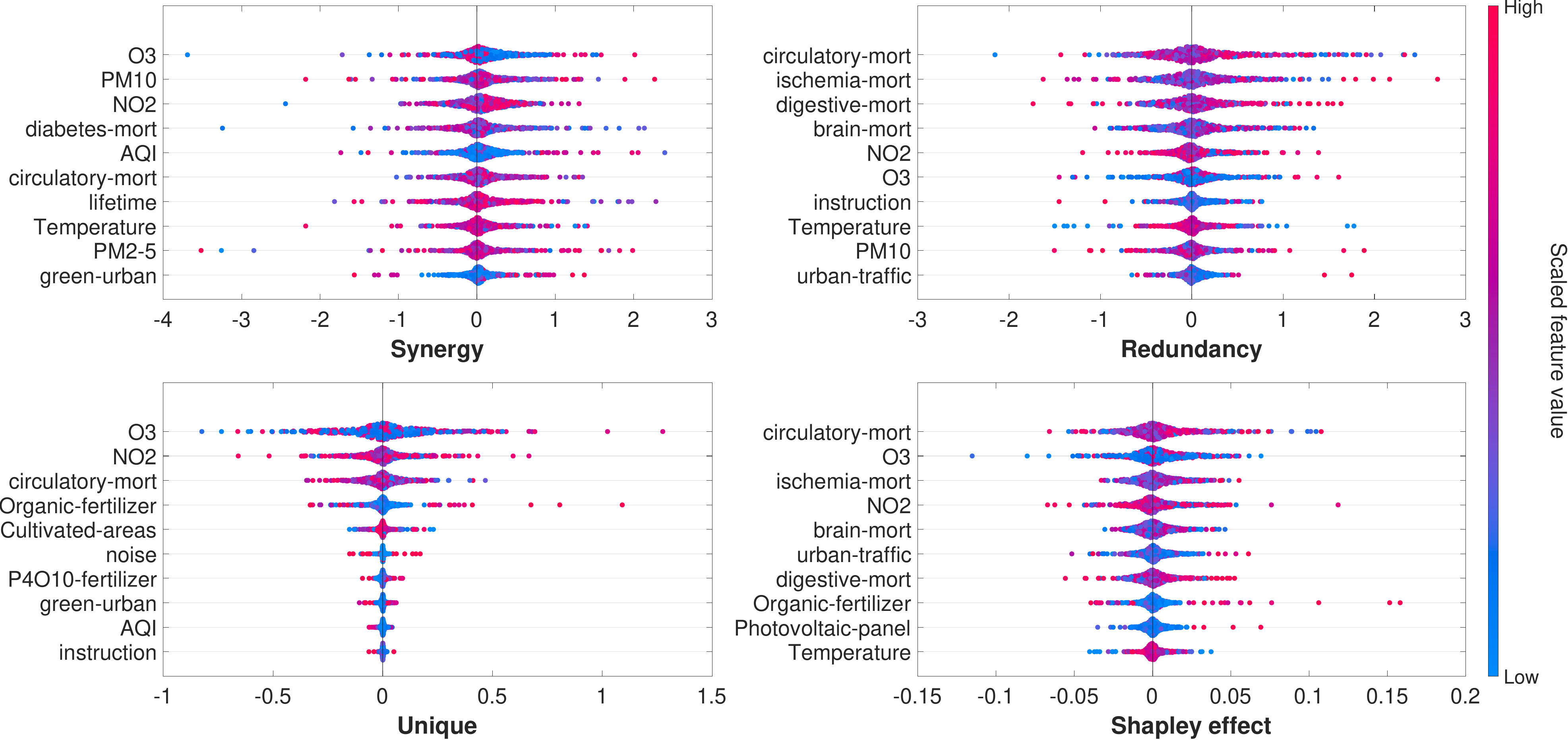}
    \caption{\textbf{Local patterns for the Alzheimer dataset. Features in each panel are ordered according to average value. Top-left panel: Local synergy of patterns. Top-right panel: Local redundancy of patterns. Bottom-left panel: Local unique scores of patterns. Bottom-right panel: Local Shapley effect.}}
    \label{fig:alz_ls}
\end{figure}

The local scores of all province patterns are depicted in figure (\ref{fig:alz_ls}) for the first ten features ranked according to their corresponding global scores (we remark that global scores are the average of local scores). As already noticed, the most synergistic features are $O_3$, $PM_{10}$, and $NO_2$ variables, which are related to air pollution. The most redundant features correspond to co-morbidity conditions (e.g. other pathologies such as diabetes-mort and brain-mortality) followed by $NO_2$ and $O_3$. Interestingly, these two air pollution factors, also provide the highest unique information, followed by the circulatory-mort variable. Air pollution and other diseases are the domains that contribute most to all local patterns. However, the unique patterns also involve the soil pollution domain. 
In figure (\ref{fig:alz_ls}) we also report the results from the first ten features ordered by Shapley effect: all the variables highlighted by Shapley effect have high rank in the local Hi-Fi decomposition. However, Shapley effect hides the specific nature of the contribution of features, redundant, synergistic or unique: the first three features according to Shapley effect are circulatory-mort, $O_3$, and ischemia-mort, local HiFi analysis put in evidence that circulatory-mort and ischemia-mort are mostly redundant with other variables, whilst $O_3$ shows unique contribution to predictability as well as synergistic contributions with $NO_2$, $PM_{10}$ and green-urban.

\begin{figure}[t!]
    \centering
    \includegraphics[width=.45\textwidth]{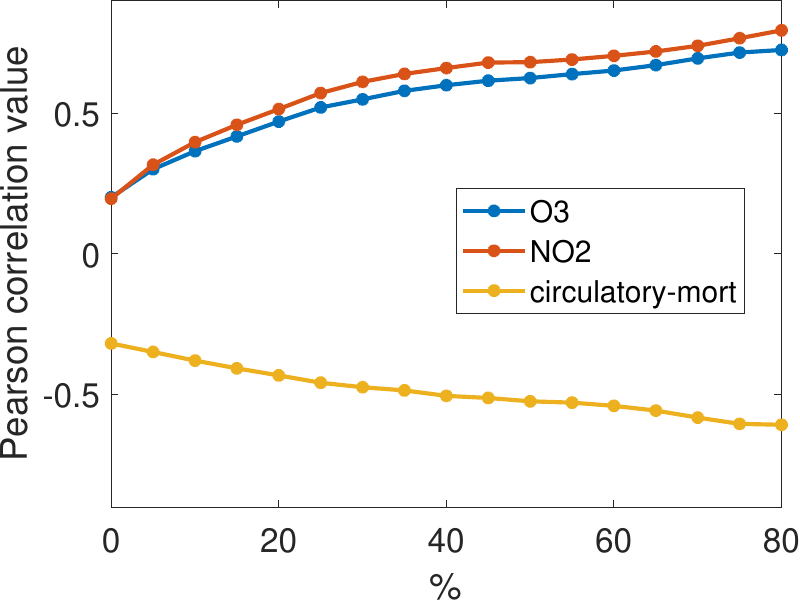}
    \caption{\textbf{Pearson correlation between the three most important  features (ranked according to global U) and AD mortality rate, as a function of the percentage of discarded patterns (the patterns with lowest U are discarded).}}
    \label{fig:alz_Uth}
\end{figure}

Figure (\ref{fig:alz_Uth}) is intended to show the meaning that unique contribution U has for each pattern. Given a pattern and an input feature, U measures how effective that feature alone was in predicting the output of that pattern. Focusing on the variables with higher global U ($O_3$, $NO_2$ and circulatory-mort), we discard the samples with lowest U by an amount (percent) which corresponds to the horizontal axis; then we plot (as a function of the percentage of discarded patterns) the Pearson correlation between features and mortality rate.
We observe that the correlations of $O_3$ and $NO_2$, with the target variable, when considering only the samples with higher values of U. On the other hand, the correlation between circulatory-mort and the output becomes more negative:
this negative association is due to the fact we are using mortality (and not incidence) rates and circulatory problems have high prevalence both in the general population and in Alzheimer's population, negative correlation thus expresses the fact that deaths are ascribed to just one of the two.  

\begin{figure}[t!]
    \centering
    \includegraphics[width=.6\textwidth]{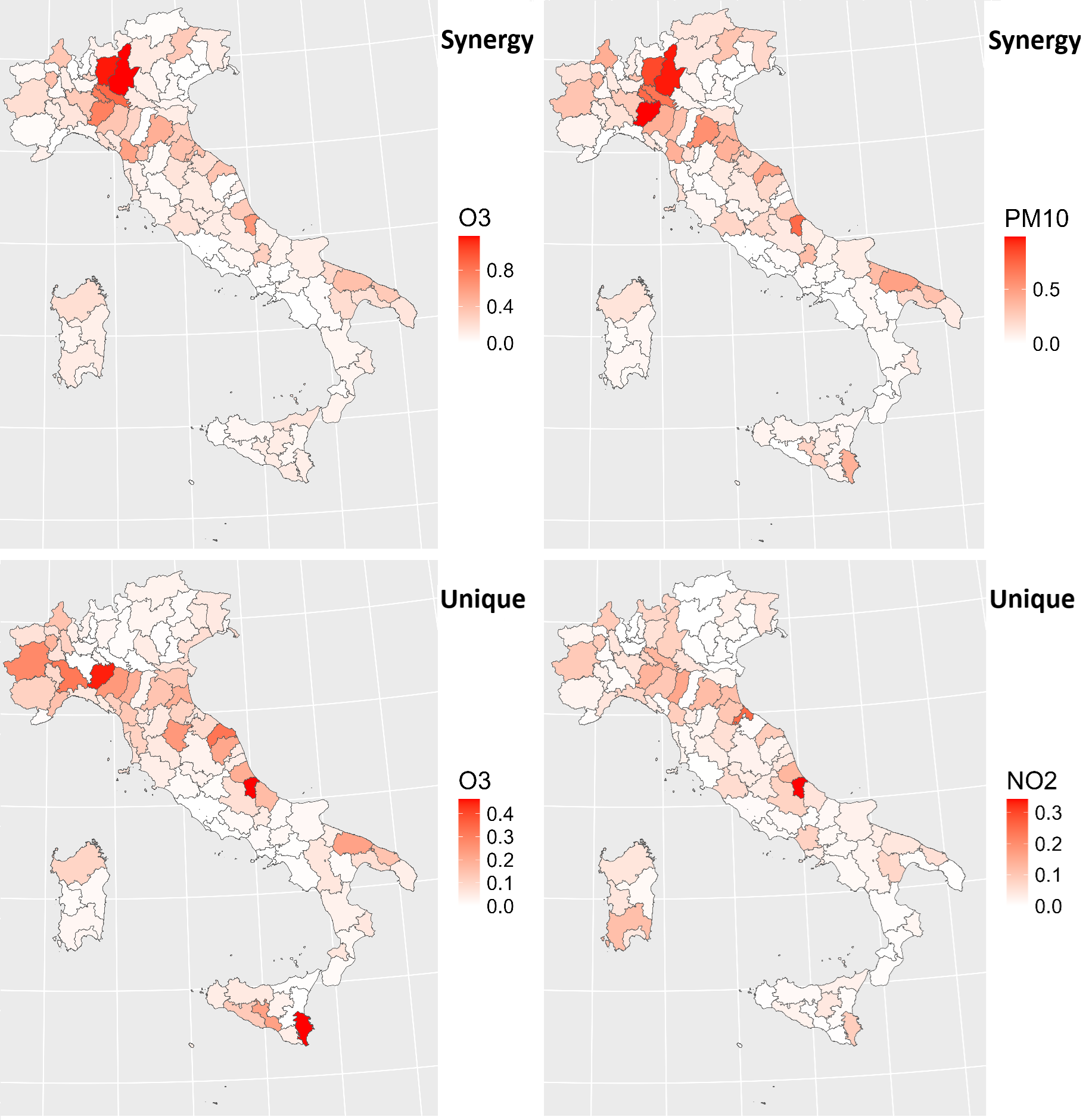}
    \caption{\textbf{Average values (over the five years) of the local synergistic (Top) and unique (Bottom) scores, for the two most important features, in each Italian province.}}
    \label{fig:ita}
\end{figure}

Figure (\ref{fig:ita}) shows the average (over the five years) of the local synergistic and unique scores for the two most important features in each Italian province. The two panels on top refer to the synergy of $O_3$ and $PM_10$, and show that synergy has higher importance than U (the two panels on bottom): pollutants features act cooperatively rather than in a pure two-body way. The provinces of Bergamo and Brescia, in Lombardy, appear to be those in which the influence of pollutants is most synergistic: note that in these two provinces the unique components of $O_3$ are very low. Concerning U: for $O_3$ Piacenza, Pescara e Siracusa show the highest local unique scores, whilst Pescara shows high unique score also for $NO_2$ pollutant. It is worth mentioning that pollutants seem to cooperate synergistically mostly in the north of Italy, whilst in the South they mostly show a pure two-body influence on AD mortality.

Summarizing, the multifactorial nature of AD is evidenced in our results, with interactions among factors belonging to different domains. The most synergistic features for predicting mortality were those related to air pollution. 
This result, together with the strong unique contributions of $NO_2$ and $O_3$, reinforces findings in the existing literature that identify air pollution as a major risk factor for Alzheimer's disease. For a discussion of potential mechanisms by which air pollution may contribute to neurodegenerative disorders, we refer the reader to \cite{alz_data}.


\section{Conclusions}

We proposed the local version of the Hi-Fi feature importance decomposition, exploiting the adaptive LOCO index to highlight high-order dependencies between features, to assign to each pattern three scores: synergistic, redundant and unique scores. 
This allows us to quantify the extent to which each feature contributes—either independently or in cooperation with others—to predicting the output for each pattern.
We demonstrate that our approach is complementary to both Shapley effects and Shapley values, as it enables a clearer disambiguation of each covariate’s role across different patterns.

We considered an interesting one-health application, i.e. the relation between air pollutants and Alzheimer's disease mortality rate. We found that air pollutants act synergistically (and in cooperation with the density of urban green areas) to improve the prediction of AD mortality: 
our findings are consistent with emerging hypotheses regarding the biological mechanisms that may explain the observed link between air pollution and Alzheimer's disease.
In our view, Local HiFi serves as a valuable tool for explainable AI (XAI) and offers a powerful approach for capturing and describing interdependencies among components in complex systems.

\begin{acknowledgments}
This research was funded by the project  “Higher-order complex systems modeling for personalized medicine”, Italian Ministry of University and Research (funded by MUR, PRIN 2022-PNRR, code P2022JAYMH, CUP: H53D23009130001) (SS). RB, AM and SS were supported by the Italian funding within the “Budget MIUR-Dipartimenti di Eccellenza 2023 - 2027” (Law 232, 11
December 2016) - Quantum Sensing and Modelling for
One-Health (QuaSiModO), CUP:H97G23000100001.
\end{acknowledgments}

\bibliographystyle{unsrt}
\bibliography{biblio}

\end{document}